%% file: main.tex
\documentclass[10pt,twocolumn,letterpaper]{article}

\usepackage{cvpr}
\usepackage{times}
\usepackage{epsfig}
\usepackage{graphicx}
\usepackage{amsmath}
\usepackage{amssymb}
\usepackage{bm}
\usepackage{booktabs}
\usepackage{tabularx}
\usepackage{changes}
\colorlet{Changes@Color}{red}
\usepackage{multirow}
\usepackage{subcaption}

\newcolumntype{P}[1]{>{\raggedright\arraybackslash}m{#1}}%
\newcolumntype{C}[1]{>{\centering\arraybackslash}m{#1}}%
\newcolumntype{R}[1]{>{\raggedleft\arraybackslash}m{#1}}%

\usepackage[pagebackref=true,breaklinks=true,letterpaper=true,colorlinks=false,bookmarks=false,allbordercolors={white}]{hyperref}

\cvprfinalcopy %

\usepackage{microtype}

\ifcvprfinal\fi

\begin{document}

\title{It's Written All Over Your Face:\\Full-Face Appearance-Based Gaze Estimation}

\author{
Xucong Zhang$^1$
\and
Yusuke Sugano$^3$\thanks{This work was mainly done while at the Max Planck Institute for Informatics}
\and
Mario Fritz$^2$
\and
Andreas Bulling$^1$\\
\and
$^1$Perceptual User Interfaces Group, $^2$Scalable Learning and Perception Group\\
Max Planck Institute for Informatics, Saarland Informatics Campus, Germany\\
$^3$Graduate School of Information Science and Technology, Osaka University, Japan\\
{\tt\small \{xczhang,mfritz,bulling\}@mpi-inf.mpg.de sugano@ist.osaka-u.ac.jp}
}

\maketitle
\thispagestyle{empty}

\begin{abstract}
\input{abstract.tex}
\end{abstract}

\input{introduction.tex}
\input{relatedwork.tex}
\input{method.tex}

\input{fullface.tex}

\input{experiments.tex}

\input{analysis.tex}
\input{conclusion.tex}

\section{Acknowledgements}

This work was partly funded by the Cluster of Excellence on Multimodal Computing and Interaction (MMCI) at Saarland University, Germany, and JST CREST Research Grant (JPMJCR14E1), Japan.

{\small
\bibliographystyle{ieee_fullname}
\bibliography{reference}
}

\end{document}

%% file: abstract.tex
Eye gaze is an important non-verbal cue for human affect analysis.
Recent gaze estimation work indicated that information from the full face region can benefit performance.
Pushing this idea further, we propose an appearance-based method that, in contrast to a long-standing line of work in computer vision, only takes the full face image as input.
Our method encodes the face image using a convolutional neural network with spatial weights applied on the feature maps to flexibly suppress or enhance information in different facial regions.
Through extensive evaluation, we show that our full-face method significantly outperforms the state of the art for both 2D and 3D gaze estimation, achieving improvements of up to 14.3\% on MPIIGaze and 27.7\% on EYEDIAP for person-independent 3D gaze estimation.
We further show that this improvement is consistent across different illumination conditions and gaze directions and particularly pronounced for the most challenging extreme head poses.

%% file: introduction.tex
\section{Introduction}

A large number of works in computer vision have studied the problem of estimating human eye gaze~\cite{hansen2010eye} given its importance for different applications, such as human-robot interaction~\cite{mutlu2009footing}, affective computing~\cite{d2012gaze}, and social signal processing~\cite{vinciarelli2008social}.
While early methods typically required settings in which lighting conditions or head pose could be controlled~\cite{lu2014adaptive,pomerleau1993non,tan2002appearance,williams2006sparse}, latest appearance-based methods using convolutional neural networks (CNN) have paved the way for gaze estimation in everyday settings that are characterised by significant amount of lighting and appearance variation~\cite{zhang2015appearance}.
Despite these advances, previous appearance-based methods have only used image information encoded from one or both eyes.

\begin{figure}[t]
\center
\includegraphics[width=0.95\columnwidth]{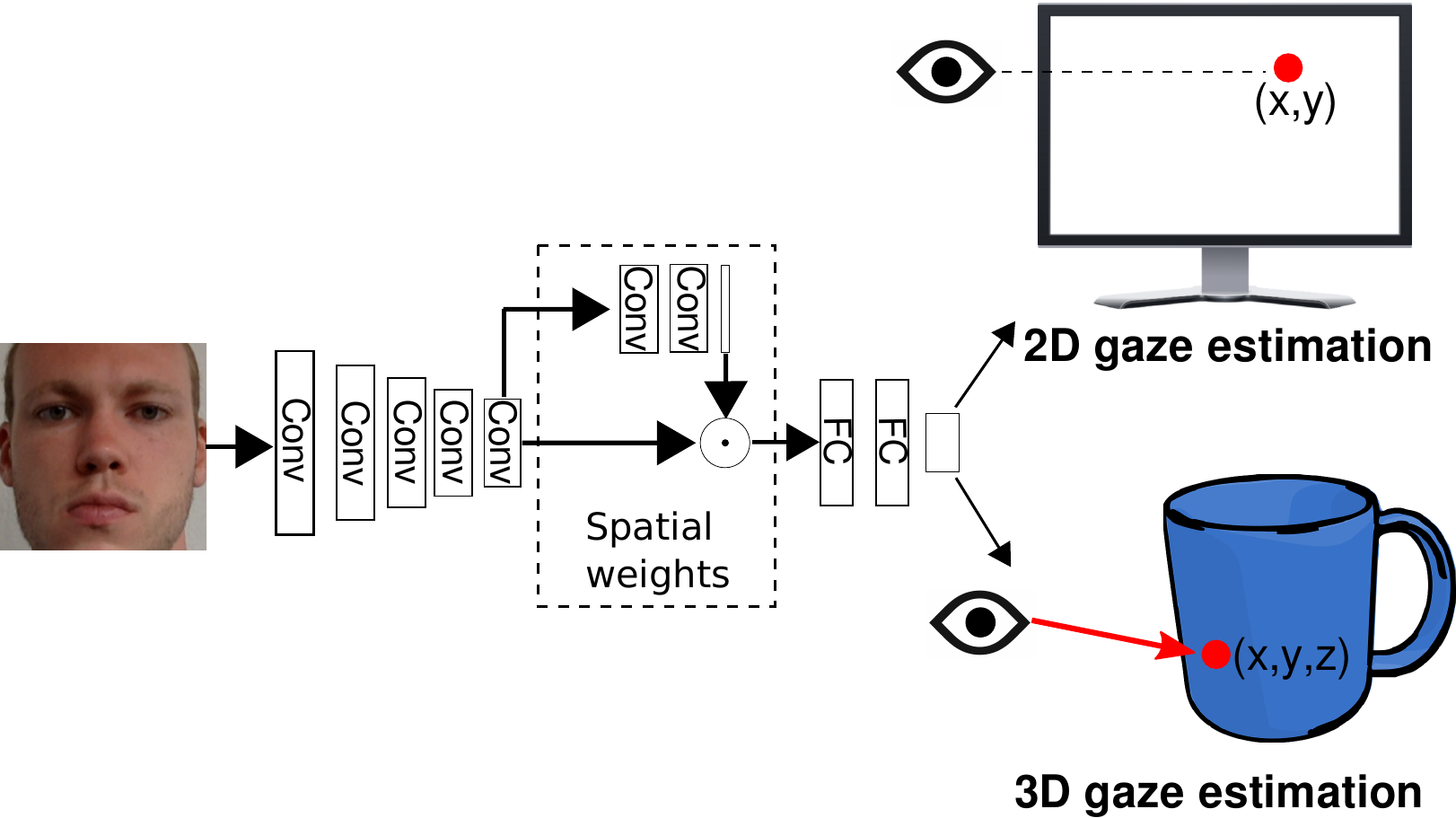}
\caption{Overview of the proposed full face appearance-based gaze estimation pipeline. Our method only takes the face image as input and performs 2D and 3D gaze estimation using a convolutional neural network with spatial weights applied on the feature maps.}
\label{fig:pipeline}
\end{figure}

Recent results by Krafka et al.\ indicated that a multi-region CNN architecture that takes both eye and face images as input can benefit gaze estimation performance~\cite{krafka2016eye}.
While, intuitively, human gaze is closely linked to eyeball pose and eye images should therefore be sufficient to estimate gaze direction, it is indeed conceivable that especially machine learning-based methods can leverage additional information from other facial regions.
These regions could, for example, encode head pose or illumination-specific information across larger image areas than those available in the eye region.
However, it is still an open question whether a (more efficient and elegant) face-only approach can work, which facial regions are most important for such a full-face appearance-based method, and whether current deep architectures can encode the information in these regions.
In addition, the gaze estimation task in~\cite{krafka2016eye} was limited to a simple 2D screen mapping and the potential of the full-face approach for 3D gaze estimation thus remains unclear.

The goal of this work is to shed light on these questions by providing a detailed analysis of the potential of the full-face approach for 2D and 3D appearance-based gaze estimation (see Fig. \ref{fig:pipeline}).
The specific contributions of this work are two-fold.
First, we propose a full-face CNN architecture for gaze estimation that, in stark contrast to a long-standing tradition in gaze estimation, takes the full face image as input and directly regresses to 2D or 3D gaze estimates.
We quantitatively compare our full-face method with existing eye-only~\cite{zhang2015appearance} and multi-region~\cite{krafka2016eye} methods and show that it can achieve a person-independent 3D gaze estimation accuracy of 4.8$^\circ$ on the challenging MPIIGaze dataset, thereby improving by 14.3\% over the state of the art.
Second, we propose a {\em spatial weights} mechanism to efficiently encode information about different regions of the full face into a standard CNN architecture.
The mechanism learns spatial weights on the activation maps of the convolutional layers, reflecting that the information contained in different facial regions.
Through further quantitative and qualitative evaluations we show that the proposed spatial weights network facilitates the learning of estimators that are robust to significant variation in illumination conditions as well as head pose and gaze directions available in current datasets.

%% file: relatedwork.tex
\section{Related Work}

Our work is related to previous works on appearance-based gaze estimation for both the 2D and 3D gaze estimation task, in particular recent multi-region methods, and means to encode spatial information in CNNs.

\vspace{-1em}
\paragraph{Appearance-Based Gaze Estimation}

Gaze estimation methods are typically categorised as either model-based or appearance-based.
While model-based methods estimate gaze direction using geometric models of the eyes and face~\cite{chen20083d,valenti2012combining,wood2014eyetab}, appearance-based methods directly regress from eye images to gaze direction. 
Early appearance-based methods assumed a fixed head pose and training data for each user~\cite{pomerleau1993non,tan2002appearance,williams2006sparse}. 
Later works focused on pose-independent gaze estimation either from monocular RGB~\cite{lu2014learning,sugano2015appearance} or depth images~\cite{funes2016gaze} but still required person-specific training.
A promising direction to achieve pose- and person-independence are learning-based methods but these require large amounts of labelled training data~\cite{krafka2016eye,sugano2014learning,zhang2015appearance}.
Consequently, recent years have seen an increasing number of gaze estimation datasets collected in everyday settings~\cite{he2015omeg,smith2013gaze}, including some at large scale~\cite{krafka2016eye,zhang2015appearance}, or consisting of synthetic data~\cite{sugano2014learning,wood16_etra,wood2015rendering}.
In this work, we also focus on this most challenging pose- and person-independent gaze estimation task using a leave-one-person-out cross-validation scheme.

\vspace{-1em}
\paragraph{2D vs. 3D Gaze Estimation}

Appearance-based gaze estimation methods can be further categorised depending on whether the regression target is in 2D or 3D.
Early works assumed a fixed head pose of the target person~\cite{pomerleau1993non,tan2002appearance,valenti2012combining,williams2006sparse}, and consequently focused on the 2D gaze estimation task where the estimator is trained to output on-screen gaze locations.
While more recent methods use 3D head pose~\cite{lu2015gaze,sugano2015appearance} or size and location of the face bounding box~\cite{krafka2016eye} to allow for free head movement, they still formulate the task as a direct mapping to 2D on-screen gaze locations.
The underlying assumption behind these 2D approaches is that the target screen plane is fixed in the camera coordinate system.
Therefore it does not allow for free camera movement after training, which can be a practical limitation especially to learning-based person-independent estimators.

In contrast, in 3D gaze estimation, the estimator is trained to output 3D gaze directions in the camera coordinate system~\cite{funes2015gaze,lu2014learning,lu2015gaze,wood2015rendering,zhang2015appearance}. 
The 3D formulation is closely related to pose- and person-independent training approaches, and the most important technical challenge is how to efficiently train estimators without requiring too much training data.
To facilitate model training, Sugano et al.\ proposed a data normalisation technique to restrict the appearance variation into a single, normalized training space~\cite{sugano2014learning}.
Although it required additional technical components, such as 3D head pose estimation, 3D methods have a technical advantage in that they can estimate gaze locations for any target object and camera setup.
Since these two approaches handle geometry information differently, the role of the full-face input can be also different between 2D and 3D approaches.

\vspace{-1em}
\paragraph{Multi-Region Gaze Estimation}

Despite these advances, most previous works used a single eye image as input to the regressor and only few considered alternative approaches, such as using two images, one of each eye~\cite{huang2015tabletgaze}, or a single image covering both eyes~\cite{he2015omeg}.
Krafka et al.\ recently presented a multi-region 2D gaze estimation method that took individual eye images, the face image, and a face grid as input~\cite{krafka2016eye}.
Their results suggested that adding the face image can be beneficial for appearance-based gaze estimation.
Our work is first to explore the potential of using information on the full face for both 2D and 3D appearance-based gaze estimation.
Pushing this idea forward, we further propose the first method that learns a gaze estimator only from the full face image in a truly end-to-end manner.

\vspace{-1em}
\paragraph{Spatial Encoding in CNNs}
Convolutional neural networks were not only successful for classification~\cite{krizhevsky2012imagenet} but also regression~\cite{Simonyan14c}, including gaze estimation~\cite{zhang2015appearance}.
Several previous works encoded spatial information more efficiently, for example by cropping sub-regions of the image~\cite{girshickICCV15fastrcnn,jaderberg2015spatial} or treating different regions on the image equally~\cite{he2014spatial}.
Tompson et al.\ used a spatial dropout before the fully connected layer to avoid overfitting during training, but the dropout extended to the entire feature maps instead of one unit~\cite{tompson2015efficient}.
We instead propose a spatial weights mechanism that encodes the weights for the different region of full face, suppress noisy and enhance the contribution from low activation regions.

%% file: method.tex
\section{Gaze Estimation Tasks}

Before detailing our model architecture for full-face appearance-based gaze estimation, we first formulate and discuss two different gaze estimation tasks: 2D and 3D gaze estimation.
A key contribution of this work is to investigate full-face appearance-based gaze estimation for both tasks.
This not only leads to a generic model architecture but also provides valuable insights into the difference and benefits gained from full-face information for both task formulations.

Although the 3D task formulation poses additional technical challenges to properly handle the complex 3D geometry, it can be applied to different device and setups without assuming a fixed camera-screen relationship.
This formulation therefore is the most general and practically most relevant.
If the application scenario can afford a fixed screen position, the 2D formulation is technically less demanding and therefore expected to show better accuracy.

\subsection{2D Gaze Estimation}

As the most straightforward strategy, the 2D gaze estimation task is formulated as a regression from the input image $\bm{I}$ to a 2-dimensional on-screen gaze location $\bm{p}$ as $\bm{p} = f(\bm{I})$, where $f$ is the regression function.
Usually $\bm{p}$ is directly defined in the coordinate system of the target screen~\cite{lu2014adaptive,sugano2015appearance,tan2002appearance,valenti2012combining} or, more generally, a virtual plane defined in the camera coordinate system~\cite{krafka2016eye}.
Since the relationship between eye appearance and gaze location depends on the position of the head, the regression function usually requires 3D head poses~\cite{valenti2012combining} or face bounding box locations~\cite{huang2015tabletgaze,krafka2016eye} in addition to eye and face images.

It is important to note that, in addition to the fixed target plane, another important assumption in this formulation is that the input image $\bm{I}$ is always taken from the same camera with fixed intrinsic parameters.
Although no prior work explicitly discussed this issue, trained regression functions cannot be directly applied to different cameras without proper treatment of the difference in projection models.

\subsection{3D Gaze Estimation}

In contrast, the 3D gaze estimation task is formulated as a regression from the input image $\bm{I}$ to a 3D gaze vector $\bm{g} = f(\bm{I})$.
Similarly as for the 2D case, the regression function $f$ typically takes the 3D head pose as an additional input.
The gaze vector $\bm{g}$ is usually defined as a unit vector originating from a 3D reference point $\bm{x}$ such as the center of the eye~\cite{funes2015gaze,lu2014learning,lu2015gaze,wood2015rendering,zhang2015appearance}.
By assuming a calibrated camera and with information on the 3D pose of the target plane, the 3D gaze vector $\bm{g}$ can be converted by projecting gaze location $\bm{p}$ into the camera coordinate system.
The gaze location $\bm{p}$ as in the 2D case can be obtained by intersecting the 3D gaze vector $\bm{g}$ with the target plane.

\vspace{-1em}
\paragraph{Image Normalization}

To both handle different camera parameters and address the task of cross-person training efficiently, Sugano et al.\ proposed a data normalization procedure for 3D appearance-based gaze estimation~\cite{sugano2014learning}.
The basic idea is to apply a perspective warp to the input image so that the estimation can be performed in a normalized space with fixed camera parameters and reference point location.
Given the input image $\bm{I}$ and the location of the reference point $\bm{x}$, the task is to compute the conversion matrix $\bm{M} = \bm{S}\bm{R}$.

$\bm{R}$ is the inverse of the rotation matrix that rotates the camera so that it looks at the reference point and so that the $x$-axes of both the camera and head coordinate systems become parallel.
The scaling matrix $\bm{S}$ is defined so that the reference point is located at a distance $d_s$ from the origin of the normalized camera coordinate system.

The conversion matrix $\bm{M}$ rotates and scales any 3D points in the input camera coordinate system to the normalized coordinate system, and the same conversion can be applied to the input image $\bm{I}$ via perspective warping using the image transformation matrix $\bm{W} = \bm{C}_s\bm{M}\bm{C}^{-1}_r$.
$\bm{C}_r$ is the projection matrix corresponding to the input image obtained from a camera calibration, and $\bm{C}_s$ is another predefined parameter that defines the camera projection matrix in the normalized space.

During training, all training images $\bm{I}$ with ground-truth gaze vectors $\bm{g}$ are normalized to or directly synthesized~\cite{sugano2014learning,wood2015rendering} in the training space, which is defined by $d_s$ and $\bm{C}_s$. 
Ground-truth gaze vectors are also normalized as $\bm{\hat g} = \bm{M}\bm{g}$, while in practice they are further converted to an angular representation (horizontal and vertical gaze direction) assuming a unit length.
At test time, test images are normalized in the same manner and their corresponding gaze vectors in the normalized space are estimated via regression function trained in the normalized space.
Estimated gaze vectors are then transformed back to the input camera coordinates by $\bm{g} = \bm{M}^{-1}\bm{\hat g}$.

%% file: fullface.tex
\section{Full-Face Gaze Estimation with a Spatial Weights CNN}
\label{sec:spatial_weights}

For both the 2D and 3D gaze estimation case, the core challenge is to learn the regression function $f$.
While a large body of work has only considered the use of the eye region for this task, we instead aim to explore the potential of extracting information from the full face.

Our hypothesis is that other regions of the face beyond the eyes contain valuable information for gaze estimation.

\begin{figure*}[t]
\center
\includegraphics[width=0.9\textwidth]{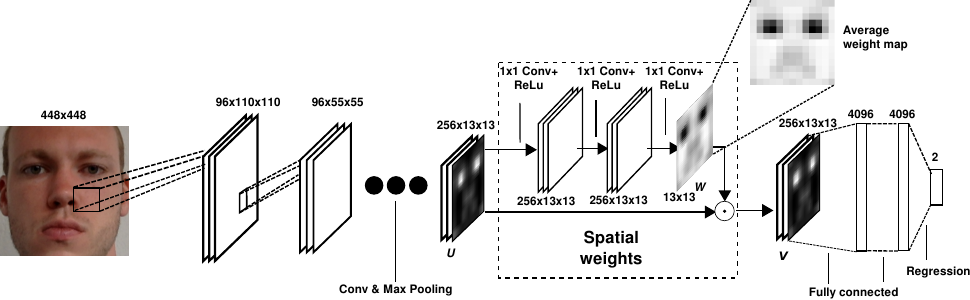}
\caption{Spatial weights CNN for full-face appearance-based gaze estimation. The input image is passed through multiple convolutional layers to generate a feature tensor $\bm{U}$. The proposed spatial weights mechanism takes $\bm{U}$ as input to generate the weight map $\bm{W}$, which is applied to $\bm{U}$ using element-wise multiplication. The output feature tensor $\bm{V}$ is fed into the following fully connected layers to -- depending on the task -- output the final 2D or 3D gaze estimate.}
\label{fig:model}
\end{figure*}

As shown in Fig.~\ref{fig:model}, to this end we propose a CNN with spatial weights (spatial weights CNN) for full-face appearance-based 2D and 3D gaze estimation.
To efficiently use the information from full-face images, we propose to use additional layers that learn spatial weights for the activation of the last convolutional layer.
The motivation behind this spatial weighting is two-fold. 
First, there could be some image regions that do not contribute to the gaze estimation task such as background regions, and activations from such regions have to be suppressed for better performance.
Second, more importantly, compared to the eye region that is expected to always contribute to the gaze estimation performance, activations from other facial regions are expected to subtle.
The role of facial appearance is also depending on various input-dependent conditions such as head pose, gaze direction and illumination, and thus have to be properly enhanced according to the input image appearance.
Although, theoretically, such differences can be learned by a normal network, we opted to introduce a mechanism that forces the network more explicitly to learn and understand that different regions of the face can have different importance for estimating gaze for a given test sample.
To implement this stronger supervision, we used the concept of the three $1 \times 1$ convolutional layers plus rectified linear unit layers from~\cite{tompson2015efficient} as a basis and adapted it to our full face gaze estimation task.
Specifically, instead of generating multiple heatmaps (one to localise each body joint) we only generated a single heatmap encoding the importance across the whole face image.
We then performed an element-wise multiplication of this weight map with the feature map of the previous convolutional layer.
An example weight map is shown in Fig.~\ref{fig:model}, averaged from all samples from the MPIIGaze dataset. 

\subsection{Spatial Weights Mechanism}

The proposed spatial weights mechanism includes three additional convolutional layers with filter size $1\times1$ followed by a rectified linear unit layer (see Fig.~\ref{fig:model}).
Given activation tensor $\bm{U}$ of size $N \times H \times W$ as input from the convolutional layer, where $N$ is the number of feature channels and $H$ and $W$ are height and width of the output, the spatial weights mechanism generates a $H \times W$ spatial weight matrix $\bm{W}$.
Weighted activation maps are obtained from element-wise multiplication of $\bm{W}$ with the original activation $\bm{U}$ with
\begin{equation}
\bm{V}_c=\bm{W}\odot\bm{U}_c,
\end{equation}
where $\bm{U}_c$ is the $c$-th channel of $\bm{U}$, and $\bm{V}_c$ corresponds to the weighted activation map of the same channel.
These maps are stacked to form the weighted activation tensor $\bm{V}$, and are fed into the next layer.
Different from the spatial dropout~\cite{tompson2015efficient}, the spatial weights mechanism weights the information continuously and keeps the information from different regions.
The same weights are applied to all feature channels, and thus the estimated weights directly correspond to the facial region in the input image.

During training, the filter weights of the first two convolutional layers are initialized randomly from a Gaussian distribution with 0 mean and 0.01, and a constant bias of 0.1.
The filter weights of the last convolutional layers are initialized randomly from a Gaussian distribution with 0 mean and 0.001 variance, and a constant bias of 1.

Gradients with respect to $\bm{U}$ and $\bm{W}$ are

\begin{equation}
\frac{\partial \bm{V}}{\partial \bm{U}}={\partial \bm{W}},
\end{equation}
and
\begin{equation}
\frac{\partial \bm{V}}{\partial \bm{W}}=\frac{1}{N}\sum_c^N{\partial \bm{U}_c}.
\end{equation}

The gradient with respect to $\bm{W}$ is normalised by the total number of the feature maps $N$, since the weight map $\bm{W}$ affects all the feature maps in $\bm{U}$ equally.

\subsection{Implementation Details}
\label{sec:model_details}

As the baseline CNN architecture we used AlexNet~\cite{krizhevsky2012imagenet} that consists of five convolutional layers and two fully connected layers.
We trained an additional linear regression layer on top of the last fully connected layer to predict the $\bm{p}$ in screen coordinates for 2D gaze estimation or normalized gaze vectors $\bm{\hat g}$ for the 3D gaze estimation task.
We used the pre-training result on the LSVRC-2010 ImageNet training set~\cite{krizhevsky2012imagenet} to initialize the five convolution layers, and fine-tuned the whole network on the MPIIGaze dataset~\cite{zhang2015appearance}.
The input image size of our networks was $448\times448$ pixels, which results in an activation $\bm{U}$ of size $256\times13\times13$ after the pooling layer of the 5-th convolutional layers.

For 2D gaze estimation, input face images were cropped according to the six facial landmark locations (four eye corners and two mouth corners). 
While in practice this is assumed to be done with face alignment methods such as~\cite{baltruvsaitis2014continuous}, in the following experiments we used dataset-provided landmark locations.
The centroid of the six landmarks was used as the center of the face, and a rectangle with a width of 1.5 times the maximum distance between landmarks was used as the face bounding box. 
The loss function was the $\ell 1$ distance between the predicted and ground-truth gaze positions in the target screen coordinate system.

For 3D gaze estimation, the reference point $\bm{x}$ was defined as the center of 3D locations of the same six facial landmarks. 
We fit the generic 3D face model provided with MPIIGaze to the landmark locations to estimate the 3D head pose.
During image normalization, we defined $d_s$ and $\bm{C}_s$ so that the input face image size became 448$\times$448 pixels.
In preliminary experiments we noticed that the additional head pose feature proposed by Zhang et al.~\cite{zhang2015appearance} did not improve the performance in the full-face case.
In this work we therefore only used image features.
The loss function was the $\ell 1$ distance between the predicted and ground-truth gaze angle vectors in the normalized space.

%% file: experiments.tex
\section{Evaluation}\label{sec:experiments}

To evaluate our architecture for the 2D and 3D gaze estimation tasks, we conducted experiments on two current gaze datasets: MPIIGaze~\cite{zhang2015appearance} and EYEDIAP~\cite{mora2014eyediap}.
For the MPIIGaze dataset, we performed a leave-one-person-out cross-validation on all 15 participants.
In order to eliminate the error caused by face alignment, we manually annotated the six facial landmarks for data normalization and image cropping. 
In the original evaluation, there were 1,500 left and 1,500 right eye samples randomly taken from each participant. 
For a direct comparison, we obtained face images corresponding to the same evaluation set and flipped the face images when they came from the right eye.
Our face patch-based setting took the middle point of face (the center of all six landmarks) as the origin of gaze direction.

For the EYEDIAP dataset, we used the screen target session for evaluation and sampled one image per 15 frames from four VGA videos of each participant.
We used head pose and eye centres annotations provided by the dataset for image normalization, and reference points were set to the midpoint of the two eye centres.
The eye images were cropped by the same way as MPIIGaze dataset.
We randomly separated the 14 participants into 5 groups and performed 5-fold cross-validation.

We compared our full-face gaze estimation method with two state-of-the-art baselines: A single eye method~\cite{zhang2015appearance} that only uses information encoded from one eye as well as a multi-region method~\cite{krafka2016eye} that takes eye images, the face image, and a face grid as input.

\vspace{-1em}
\paragraph{Single Eye}
One of the baseline methods is the state-of-the-art single eye appearance-based gaze estimation method~\cite{zhang2015appearance}, which originally used the LeNet~\cite{jia2014caffe,lecun1998gradient} architecture.
For a fair comparison, we instead used the AlexNet architecture as our proposed model (see Sec.~\ref{sec:model_details}).
Eye images were cropped by taking the center of the eye corners as the center and with the width of 1.5 times of the distance between corners, and resized to $60\times36$ pixels as proposed in~\cite{zhang2015appearance}.
In this case, each individual eye became the input to the model, and the reference point $\bm{x}$ was set to the middle of inner and outer eye corners.

\vspace{-1em}
\paragraph{iTracker}

Since neither code nor models were available, we re-implemented the iTracker architecture~\cite{krafka2016eye} according to the description provided in the paper.
Face images were cropped in the same manner as our proposed method 
and resized to $224\times224$ pixels.
Eye images were cropped by taking the middle point of the inner and outer eye corners as the image center and with the width of 1.7 times of the distance between the corners, and resized to $224\times224$ pixels. 
For the 2D gaze estimation task, we also used the face grid feature~\cite{krafka2016eye} with a size of $25\times25$ pixels.
The face grid encodes the face size and location inside the original image.
For a fair comparison with our proposed architecture, we also evaluated the model using the same AlexNet CNN architecture as {\em iTracker (AlexNet)}.
To validate the effect of the face input, we also tested the iTracker (AlexNet) architecture only taking two eye images as {\em Two eyes} model.

\subsection{2D Gaze Estimation}
\label{sec:results_2D}

\begin{figure}[t]
\center
\includegraphics[width=\columnwidth]{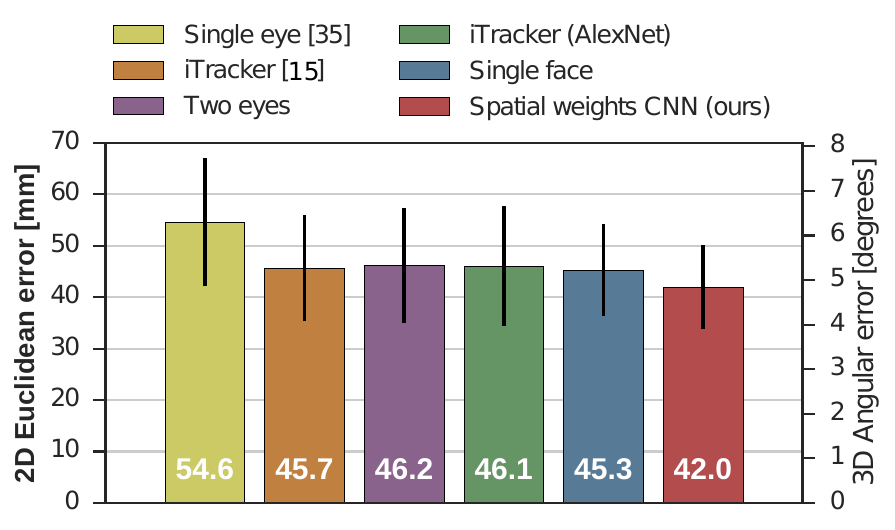}
\caption{Error for 2D gaze estimation on the MPIIGaze dataset in millimetres (Euclidean error) and degrees (angular error). The face grid was used as additional input. Error bars indicate standard deviations.}
\label{fig:performance_2d}
\end{figure}

Fig.~\ref{fig:performance_2d} summarises the results for the 2D gaze estimation task.
Each row corresponds to one method, and if not noted otherwise, the face grid feature was used in addition to the image input.
The left axis shows the Euclidean error between estimated and ground-truth gaze positions in the screen coordinate system in millimetres.
The right axis shows the corresponding angular error that was approximately calculated from the camera and monitor calibration information provided by the dataset and the same reference position for the 3D gaze estimation task. 

\begin{figure}[t]
\center
\includegraphics[width=\columnwidth]{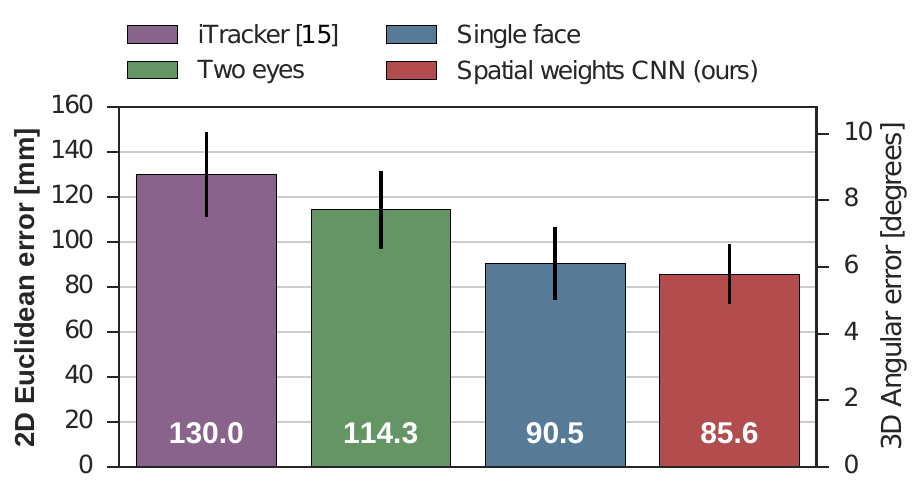}
\caption{Error for 2D gaze estimation on the EYEDIAP dataset in millimetres (Euclidean error) and degrees (angular error). Error bars indicate standard deviations.}
\label{fig:performance_2d_eyediap}
\end{figure}

As can be seen from Fig.~\ref{fig:performance_2d}, all methods that take full-face information as input significantly outperformed the single eye baseline.
The single face image model achieved a competitive result to the iTracker and the iTracker (AlexNet) models.
Performance was further improved by incorporating the proposed spatial weights network.
The proposed spatial weights network achieved a statistically significant 7.2\% performance improvement (paired t-test: $p < 0.01$) over the second best single face model.
These findings are in general mirrored for the EYEDIAP dataset shown in Fig.~\ref{fig:performance_2d_eyediap}, while the overall performance is worse most likely due to the lower resolution and the limited amount of training images.
Although the iTracker architecture performs worse than the two eyes model, our proposed model still performed the best.

\subsection{3D Gaze Estimation}
\label{sec:results_3D}

\begin{figure}[t]
\center
\includegraphics[width=\columnwidth]{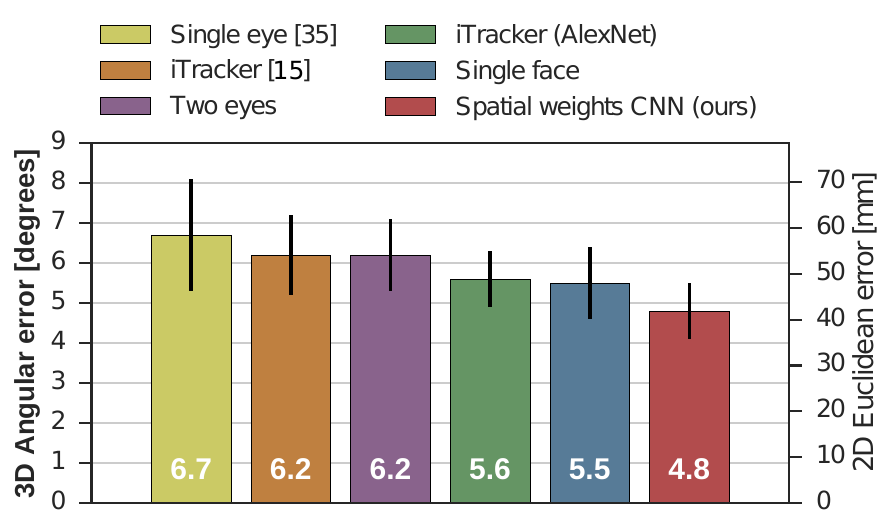}
\caption{Error for 3D gaze estimation on the MPIIGaze dataset in degrees (angular error) and millimetres (Euclidean error). Error bars indicate standard deviations.}
\label{fig:performance_3d}
\end{figure}

Fig.~\ref{fig:performance_3d} summarises the results for the 3D gaze estimation task.
The left axis shows the angular error that was directly calculated from the estimated and ground-truth 3D gaze vectors.
The right axis shows the corresponding Euclidean error that was approximated by intersecting the estimated 3D gaze vector with the screen plane.
Compared to the 2D gaze estimation task, the performance gap between iTracker and the single face model is larger (0.7 degrees). 
Since the AlexNet-based iTracker model could achieve similar performance as the single face model, the performance drop seems to be partly due to its network architecture.
Our proposed model achieved a significant performance improvement of 14.3\% (paired t-test: $p>0.01$) over iTracker, and a performance consistent with the 2D case.

\begin{figure}[t]
\center
\includegraphics[width=\columnwidth]{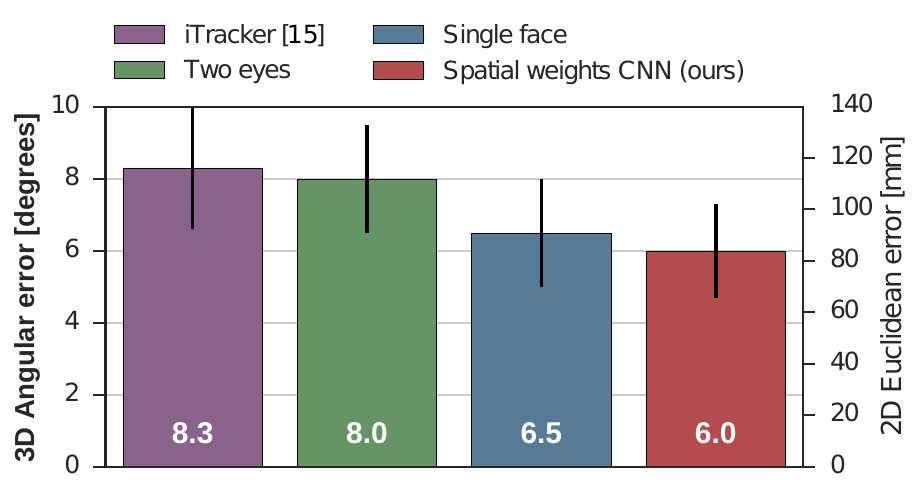}
\caption{Error for 3D gaze estimation on the EYEDIAP dataset in degrees (angular error) and millimetres (Euclidean error). Error bars indicate standard deviations.}
\label{fig:performance_3d_eyediap}
\end{figure}

As shown in Fig.~\ref{fig:performance_3d_eyediap}, the proposed model also achieved the best performance for the 3D gaze estimation task on the EYEDIAP dataset.

\subsection{Head Pose and Facial Appearance}

\begin{figure}[t]
\center
\includegraphics[width=0.95\columnwidth]{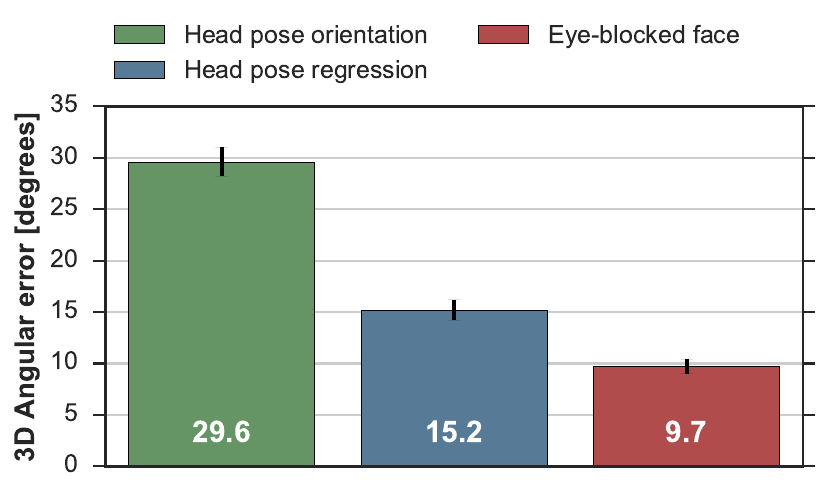}
\caption{Gaze estimation error from the different models related to head pose. The numbers are angular error for 3D gaze estimation in degrees. Error bars indicate standard deviations.}
\label{fig:performance_head_pose}
\end{figure}

One natural hypothesis about why full-face input can help the gaze estimation task is that it brings head pose information which can be a prior for inferring gaze direction.
In this section, we provide more insights on this hypothesis by comparing performance using face images {\em without} eye regions with a simple head pose-based baseline.
More specifically, using the MPIIGaze dataset, we created face images where both eye regions were blocked with a gray box according to the facial landmark annotation.
We compared the estimation performance using eye-blocked face images with: 1) a naive estimator directly treating the head pose as gaze direction, and 2) a linear regression function trained to output gaze directions from head pose input.

Angular error of these methods for the 3D estimation task are shown in Fig.~\ref{fig:performance_head_pose}.
While the error using eye-blocked face images was larger than the original single face architecture (5.5 degrees), the performance was better than baseline head pose-based estimators.
This indicates, somewhat surprisingly, that the impact of taking full-face input is larger than head pose information, and the facial appearance itself is beneficial for inferring gaze direction.

%% file: analysis.tex
\subsection{Importance of Different Facial Regions}
\label{sec:analysis}

\begin{figure}[t]
\center
\includegraphics[width=0.95\columnwidth]{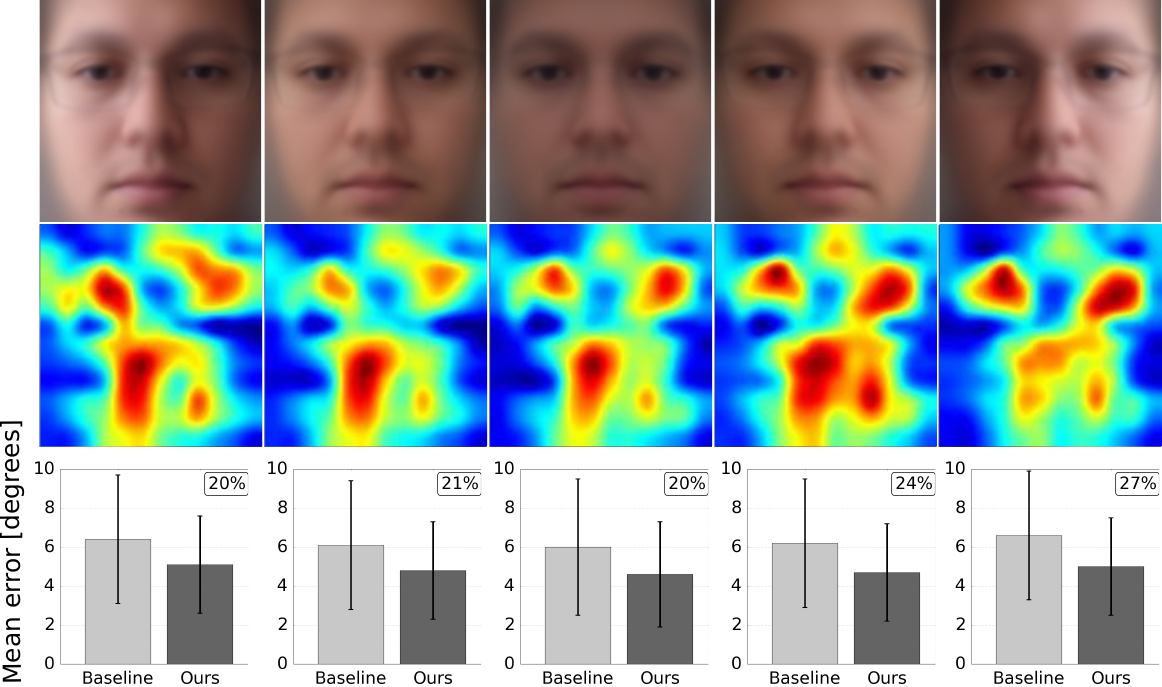}
\caption{Region importance maps and corresponding mean face patches based on a clustering of face patches according to illumination conditions for the MPIIGaze dataset: From directional light on the right side of the face (left), over frontal light (center), to directional light on the left side of the face (right). Bar plots show the estimation error for the two eye model (baseline) and the proposed spatial weights CNN (ours), and the performance gain in percent in the top right corner. Error bars indicate standard deviations.}
\label{fig:heatmap_illumination}
\end{figure}

To further analyse how different facial regions contribute to the overall performance, we generated region importance maps of the full-face model
with respect to different factors for 3D gaze estimation. 
As proposed in~\cite{zeiler2014visualizing}, region importance maps were generated by evaluating estimation error after masking parts of the input image.
Specifically, given the $448 \times 448$ input face image, we used a grey-coloured mask with a size of $64 \times 64$ pixels and moved this mask over the whole image in a sliding window fashion with a $32$ pixel stride.
The per-image region importance maps were obtained by smoothing the obtained $64 \times 64$ error distribution with a box filter.
The larger the resulting drop in gaze estimation accuracy the higher the importance of that region of the face.
Individual face images and their importance maps were then aligned by warping the whole image using three facial landmark locations (centres of both eye corners and mouth corners). 
Finally, mean face patches and mean region importance maps were computed by averaging over all images.
To illustrate the effect of the face image input, we compare these region importance maps with a quantitative performance comparison between two eyes ({\em Baseline}) and our proposed full-face model ({\em Ours}).

\vspace{-1em}
\paragraph{Illumination Conditions}
The original MPIIGaze paper characterised the dataset with respect to different illumination conditions as well as gaze ranges~\cite{zhang2015appearance}.
We therefore first explored whether and which facial regions encode information on these illumination conditions.
As in the original paper, we used the difference in mean intensity values of the right and left half of the face as a proxy to infer directional light.
We clustered all $15 \times 3,000$ images according to the illumination difference using $k$-means clustering, and computed the mean face image and mean importance map for each cluster.
Fig.~\ref{fig:heatmap_illumination} shows resulting sample region importance maps with respect to illumination conditions.
As can be seen from the figure, under strong directional lighting (leftmost and rightmost example), more widespread regions around the eyes are required on the brighter side of the face.
The proposed method consistently performed better than the two eye model over all lighting conditions.

\vspace{-1em}
\paragraph{Gaze Directions}
\begin{figure}[t]
\center
\includegraphics[width=0.95\columnwidth]{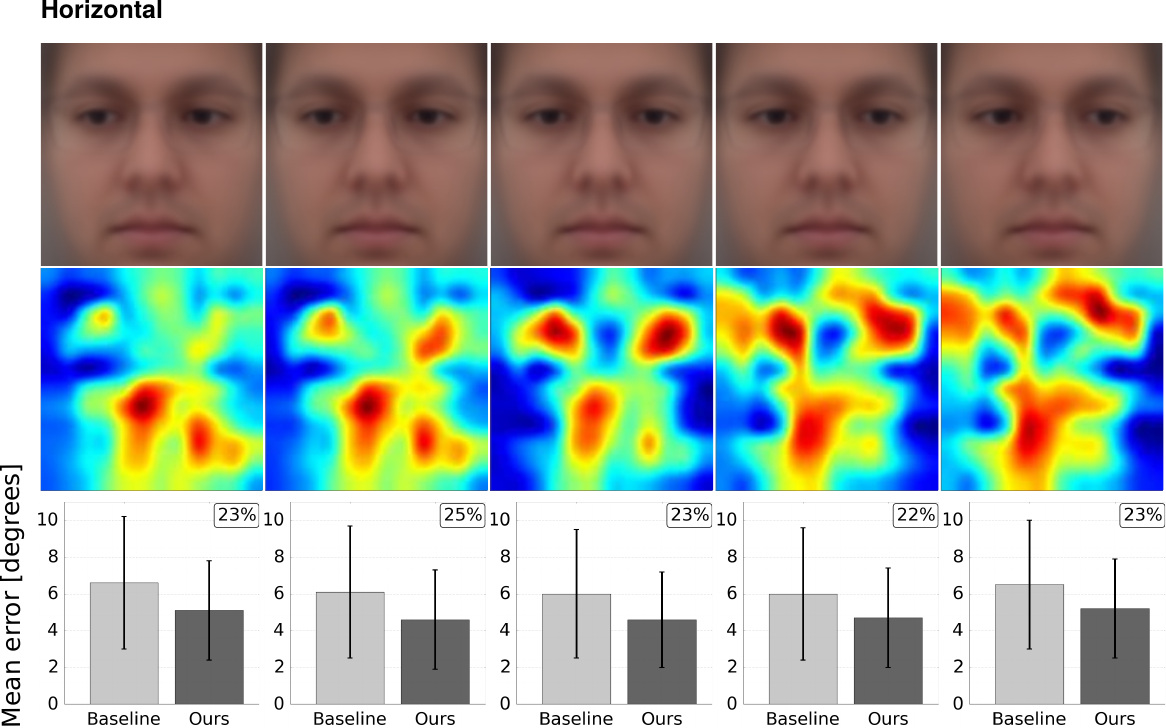}
\par
\vspace{0.2cm}
\includegraphics[width=0.95\columnwidth]{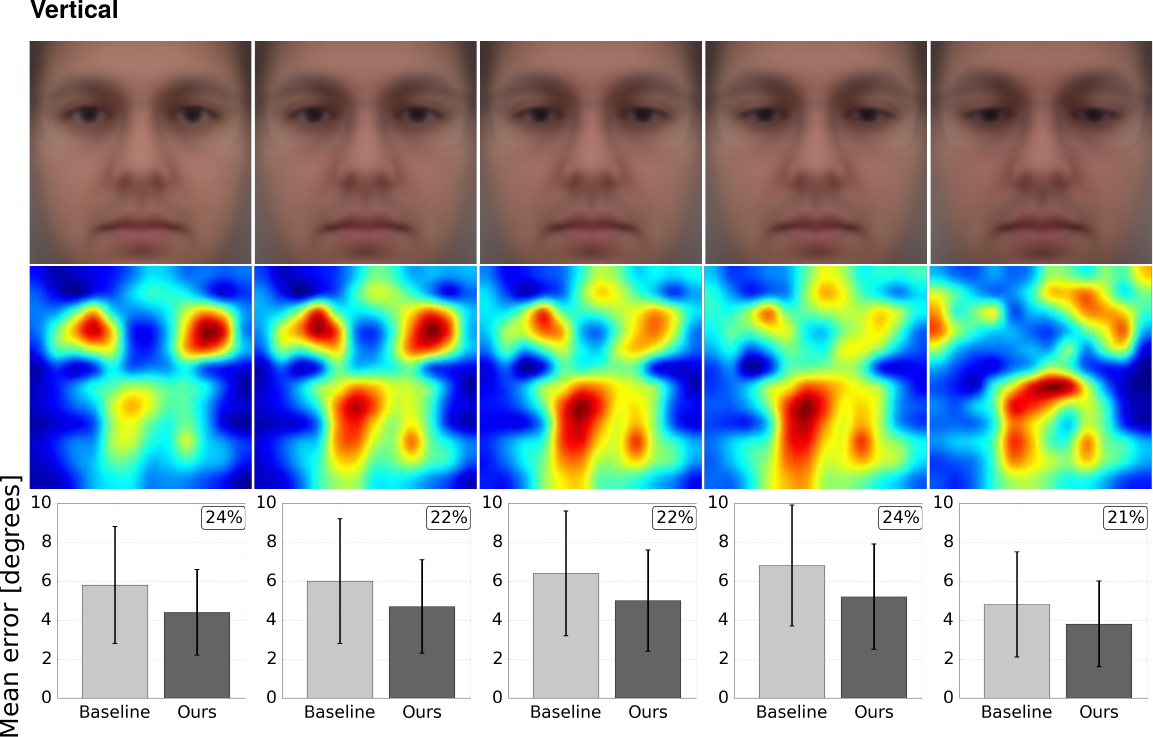}
\caption{Region importance maps and corresponding mean face patches based on a clustering of images according to ground-truth horizontal (top) and vertical (bottom) gaze direction for the MPIIGaze dataset. Bar plots show the estimation error in the same manner as in Fig.~\ref{fig:heatmap_illumination}.}
\label{fig:heatmap_gaze}
\end{figure}
Another factor that potentially influences the importance of different facial regions is the gaze direction.
We therefore clustered images according to gaze direction in the same manner as before.
The top two rows of Fig.~\ref{fig:heatmap_gaze} show the corresponding region importance maps depending on horizontal gaze direction while the bottom two rows show maps depending on vertical gaze direction.
As shown, different parts of the face become important depending on the gaze direction to be inferred.
The eye region is most important if the gaze direction is straight ahead while the model puts higher importance on other regions if the gaze direction becomes more extreme.

\vspace{-1em}
\paragraph{Head Pose}

\begin{figure}[t]
\center
\includegraphics[width=\columnwidth]{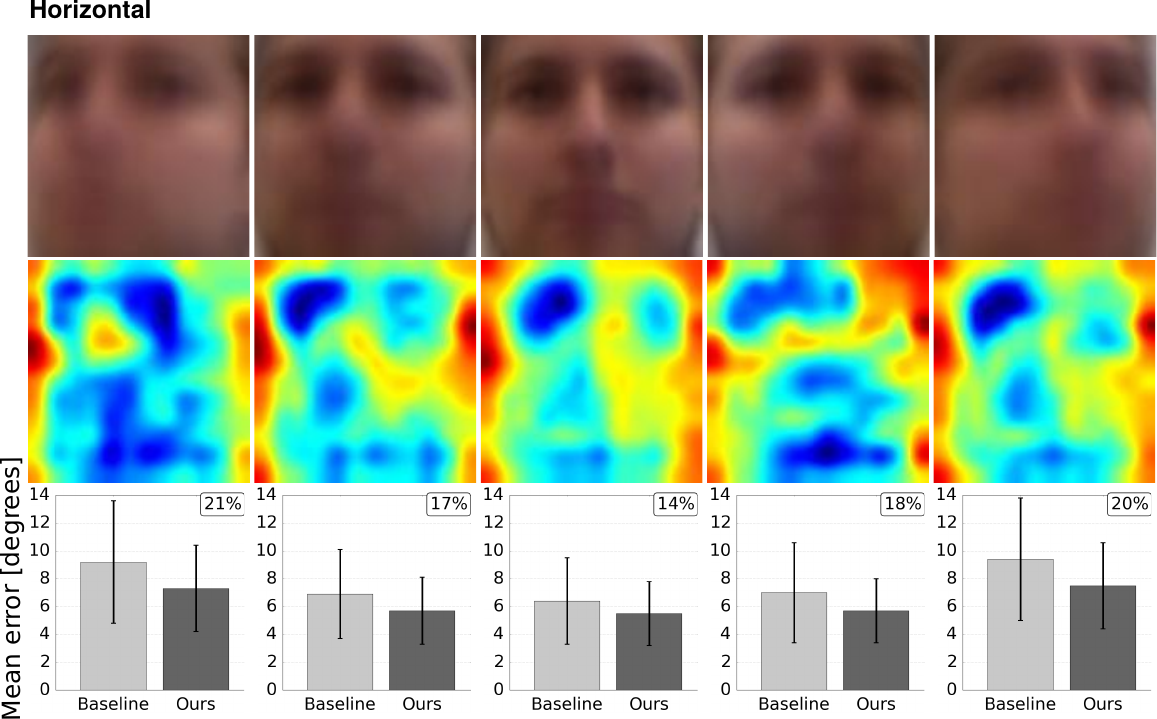}
\par
\vspace{0.2cm}
\includegraphics[width=\columnwidth]{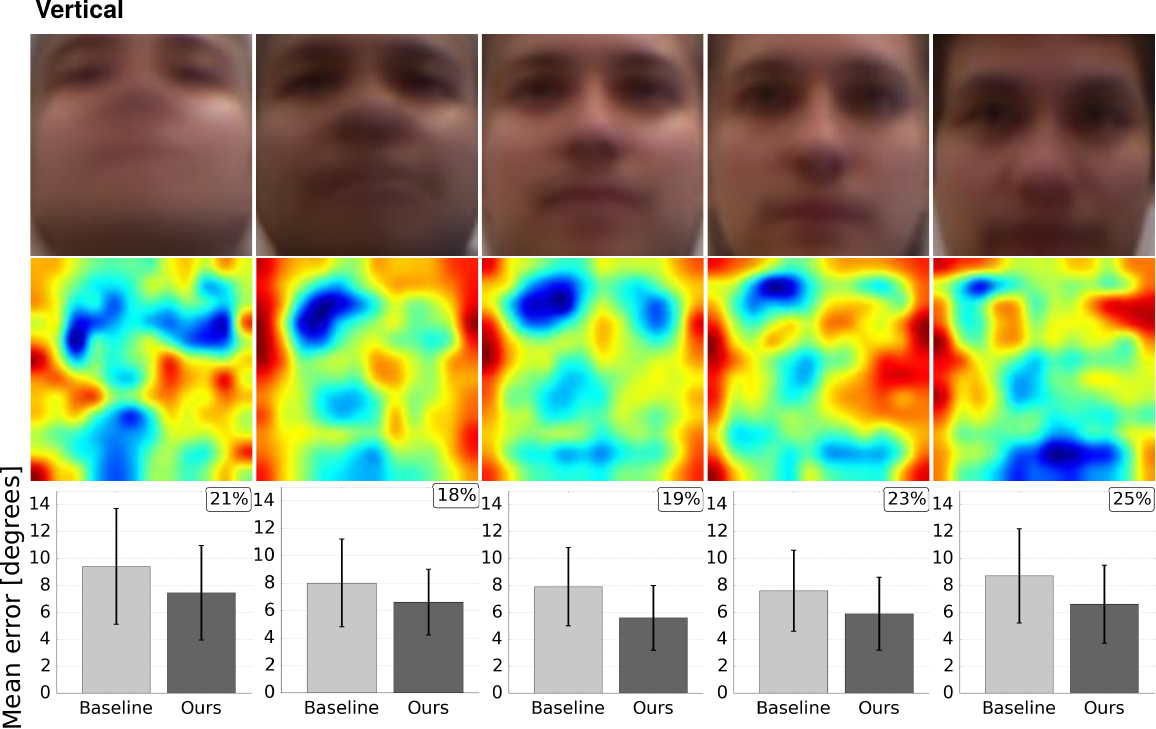}
\caption{Region importance maps based on a clustering of images according to ground-truth horizontal (top) and vertical (bottom) head pose for the EYEDIAP dataset. Bar plots show the estimation error in the same manner as in Fig.~\ref{fig:heatmap_illumination}.}
\label{fig:heatmap_headpose}
\end{figure}

While the head pose range in MPIIGaze is limited due to the recording setting, the EYEDIAP dataset contains a wide head pose range.

We therefore finally clustered images in EYEDIAP according to head pose in the same manner as before.
The top two rows of Fig.~\ref{fig:heatmap_headpose} show the corresponding region importance maps depending on horizontal head pose while the bottom two rows show maps depending on vertical head pose.
In these cases, it can be clearly seen that the full-face input is particularly beneficial to improving estimation performance for extreme head poses.
Non-eye facial regions also have in general higher importance compared to MPIIGaze, which indicates the benefit of using full-face input for low-resolution images.

%% file: conclusion.tex
\section{Conclusion}

In this work we studied full-face appearance-based gaze estimation and proposed a spatial weights CNN method that leveraged information from the full face.
We demonstrated that, compared to current eye-only and multi-region methods, our method is more robust to facial appearance variation caused by extreme head pose and gaze directions as well as illumination.
Our method achieved an accuracy of $4.8^\circ$ and $6.0^\circ$ for person-independent 3D gaze estimation on the challenging in-the-wild MPIIGaze and EYEDIAP datasets, respectively -- a significant improvement of 14.3\% and 27.7\% over the state of the art.
We believe that full-face appearance-based gaze estimation leans itself closely to related computer vision tasks, such as face and facial feature detection, facial expression analysis, or head pose estimation.
This work therefore points towards future learning-based methods that address multiple of these tasks jointly.

%% file: main.bbl
\begin{thebibliography}{10}\itemsep=-1pt

\bibitem{baltruvsaitis2014continuous}
Tadas Baltru\v{s}aitis, Peter Robinson, and Louis-Philippe Morency.
\newblock Continuous conditional neural fields for structured regression.
\newblock In {\em Proc. European Conf. Computer Vision}, pages 593--608, 2014.

\bibitem{pomerleau1993non}
Shumeet Baluja and Dean Pomerleau.
\newblock Non-intrusive gaze tracking using artificial neural networks.
\newblock Technical report, DTIC Document, 1994.

\bibitem{chen20083d}
Jixu Chen and Qiang Ji.
\newblock 3d gaze estimation with a single camera without ir illumination.
\newblock In {\em Proc. IEEE Int. Conf. Pattern Recognition}, pages 1--4, 2008.

\bibitem{d2012gaze}
Sidney D'Mello, Andrew Olney, Claire Williams, and Patrick Hays.
\newblock Gaze tutor: A gaze-reactive intelligent tutoring system.
\newblock {\em International Journal of human-computer studies},
  70(5):377--398, 2012.

\bibitem{mora2014eyediap}
Kenneth~Alberto Funes~Mora, Florent Monay, and Jean-Marc Odobez.
\newblock {EYEDIAP}: A database for the development and evaluation of gaze
  estimation algorithms from rgb and rgb-d cameras.
\newblock In {\em Proc. ACM Symp. on Eye Tracking Research and Applications},
  pages 255--258, 2014.

\bibitem{funes2015gaze}
Kenneth~Alberto Funes~Mora and Jean-Marc Odobez.
\newblock Geometric generative gaze estimation (g3e) for remote rgb-d cameras.
\newblock In {\em Proc. IEEE Conf. Computer Vision and Pattern Recognition},
  pages 1773--1780, 2014.

\bibitem{funes2016gaze}
Kenneth~A Funes-Mora and Jean-Marc Odobez.
\newblock Gaze estimation in the 3d space using rgb-d sensors.
\newblock {\em International Journal of Computer Vision}, 118(2):194--216,
  2016.

\bibitem{girshickICCV15fastrcnn}
Ross Girshick.
\newblock Fast r-cnn.
\newblock In {\em Proceedings of the IEEE international conference on computer
  vision}, pages 1440--1448, 2015.

\bibitem{hansen2010eye}
Dan~Witzner Hansen and Qiang Ji.
\newblock In the eye of the beholder: A survey of models for eyes and gaze.
\newblock {\em IEEE Trans. Pattern Analysis and Machine Intelligence},
  32(3):478--500, 2010.

\bibitem{he2014spatial}
Kaiming He, Xiangyu Zhang, Shaoqing Ren, and Jian Sun.
\newblock Spatial pyramid pooling in deep convolutional networks for visual
  recognition.
\newblock {\em IEEE transactions on pattern analysis and machine intelligence},
  37(9):1904--1916, 2015.

\bibitem{he2015omeg}
Qiuhai He, Xiaopeng Hong, Xiujuan Chai, Jukka Holappa, Guoying Zhao, Xilin
  Chen, and Matti Pietik{\"a}inen.
\newblock Omeg: Oulu multi-pose eye gaze dataset.
\newblock In {\em Image Analysis}, pages 418--427. 2015.

\bibitem{huang2015tabletgaze}
Qiong Huang, Ashok Veeraraghavan, and Ashutosh Sabharwal.
\newblock Tabletgaze: Unconstrained appearance-based gaze estimation in mobile
  tablets, 2015.

\bibitem{jaderberg2015spatial}
Max Jaderberg, Karen Simonyan, Andrew Zisserman, et~al.
\newblock Spatial transformer networks.
\newblock {\em Advances in neural information processing systems}, 28, 2015.

\bibitem{jia2014caffe}
Yangqing Jia, Evan Shelhamer, Jeff Donahue, Sergey Karayev, Jonathan Long, Ross
  Girshick, Sergio Guadarrama, and Trevor Darrell.
\newblock Caffe: Convolutional architecture for fast feature embedding.
\newblock In {\em Proc. Int. Conf. Multimedia}, pages 675--678, 2014.

\bibitem{krafka2016eye}
Kyle Krafka, Aditya Khosla, Petr Kellnhofer, Harini Kannan, Suchendra
  Bhandarkar, Wojciech Matusik, and Antonio Torralba.
\newblock Eye tracking for everyone.
\newblock In {\em Proc. IEEE Conf. Computer Vision and Pattern Recognition},
  pages 2176--2184, 2016.

\bibitem{krizhevsky2012imagenet}
Alex Krizhevsky, Ilya Sutskever, and Geoffrey~E Hinton.
\newblock Imagenet classification with deep convolutional neural networks.
\newblock In {\em Proc. NIPS}, pages 1097--1105, 2012.

\bibitem{lecun1998gradient}
Yann LeCun, L{\'e}on Bottou, Yoshua Bengio, and Patrick Haffner.
\newblock Gradient-based learning applied to document recognition.
\newblock {\em Proc. of the IEEE}, 86(11):2278--2324, 1998.

\bibitem{lu2014learning}
Feng Lu, Takahiro Okabe, Yusuke Sugano, and Yoichi Sato.
\newblock Learning gaze biases with head motion for head pose-free gaze
  estimation.
\newblock {\em Image and Vision Computing}, 32(3):169 -- 179, 2014.

\bibitem{lu2014adaptive}
F. Lu, Y. Sugano, T. Okabe, and Y. Sato.
\newblock Adaptive linear regression for appearance-based gaze estimation.
\newblock {\em IEEE Trans. Pattern Analysis and Machine Intelligence},
  36(10):2033--2046, 2014.

\bibitem{lu2015gaze}
Feng Lu, Yusuke Sugano, Takahiro Okabe, and Yoichi Sato.
\newblock Gaze estimation from eye appearance: A head pose-free method via eye
  image synthesis.
\newblock {\em IEEE Transactions on Image Processing}, 24(11):3680--3693, 2015.

\bibitem{mutlu2009footing}
Bilge Mutlu, Toshiyuki Shiwa, Takayuki Kanda, Hiroshi Ishiguro, and Norihiro
  Hagita.
\newblock Footing in human-robot conversations: how robots might shape
  participant roles using gaze cues.
\newblock In {\em Proceedings of the 4th ACM/IEEE international conference on
  Human robot interaction}, pages 61--68, 2009.

\bibitem{Simonyan14c}
K. Simonyan and A. Zisserman.
\newblock Very deep convolutional networks for large-scale image recognition.
\newblock In {\em Int. Conf. Learning Representations}, 2015.

\bibitem{smith2013gaze}
Brian~A Smith, Qi Yin, Steven~K Feiner, and Shree~K Nayar.
\newblock Gaze locking: passive eye contact detection for human-object
  interaction.
\newblock In {\em Proc. ACM Symp. on User Interface Software and Technology},
  pages 271--280, 2013.

\bibitem{sugano2014learning}
Yusuke Sugano, Yasuyuki Matsushita, and Yoichi Sato.
\newblock Learning-by-synthesis for appearance-based 3d gaze estimation.
\newblock In {\em Proc. IEEE Conf. Computer Vision and Pattern Recognition},
  pages 1821--1828, 2014.

\bibitem{sugano2015appearance}
Yusuke Sugano, Yasuyuki Matsushita, Yoichi Sato, and Hideki Koike.
\newblock Appearance-based gaze estimation with online calibration from mouse
  operations.
\newblock {\em Trans. Human-Machine Systems}, 45(6):750--760, 2015.

\bibitem{tan2002appearance}
Kar-Han Tan, David~J Kriegman, and Narendra Ahuja.
\newblock Appearance-based eye gaze estimation.
\newblock In {\em Proc. IEEE Workshop Applications of Computer Vision}, pages
  191--195, 2002.

\bibitem{tompson2015efficient}
Jonathan Tompson, Ross Goroshin, Arjun Jain, Yann LeCun, and Christoph Bregler.
\newblock Efficient object localization using convolutional networks.
\newblock In {\em Proceedings of the IEEE conference on computer vision and
  pattern recognition}, pages 648--656, 2015.

\bibitem{valenti2012combining}
Roberto Valenti, Nicu Sebe, and Theo Gevers.
\newblock Combining head pose and eye location information for gaze estimation.
\newblock {\em IEEE Trans. Image Processing}, 21(2):802--815, 2012.

\bibitem{vinciarelli2008social}
Alessandro Vinciarelli, Maja Pantic, Herv{\'e} Bourlard, and Alex Pentland.
\newblock Social signal processing: state-of-the-art and future perspectives of
  an emerging domain.
\newblock In {\em Proceedings of the 16th ACM international conference on
  Multimedia}, pages 1061--1070, 2008.

\bibitem{williams2006sparse}
Oliver Williams, Andrew Blake, and Roberto Cipolla.
\newblock Sparse and semi-supervised visual mapping with the {S}\^{} 3{GP}.
\newblock In {\em Proc. IEEE Conf. Computer Vision and Pattern Recognition},
  pages 230--237, 2006.

\bibitem{wood16_etra}
Erroll Wood, Tadas Baltru{\v{s}}aitis, Louis-Philippe Morency, Peter Robinson,
  and Andreas Bulling.
\newblock Learning an appearance-based gaze estimator from one million
  synthesised images.
\newblock In {\em Proc. ACM Symp. Eye Tracking Research Applications}, pages
  131--138, 2016.

\bibitem{wood2015rendering}
Erroll Wood, Tadas Baltrusaitis, Xucong Zhang, Yusuke Sugano, Peter Robinson,
  and Andreas Bulling.
\newblock Rendering of eyes for eye-shape registration and gaze estimation.
\newblock In {\em Proc. IEEE Int. Conf. Computer Vision}, 2015.

\bibitem{wood2014eyetab}
Erroll Wood and Andreas Bulling.
\newblock Eyetab: Model-based gaze estimation on unmodified tablet computers.
\newblock In {\em Proc. ACM Symp. on Eye Tracking Research and Applications},
  pages 207--210, 2014.

\bibitem{zeiler2014visualizing}
Matthew~D Zeiler and Rob Fergus.
\newblock Visualizing and understanding convolutional networks.
\newblock In {\em European conference on computer vision}, pages 818--833.
  Springer, 2014.

\bibitem{zhang2015appearance}
Xucong Zhang, Yusuke Sugano, Mario Fritz, and Andreas Bulling.
\newblock Appearance-based gaze estimation in the wild.
\newblock In {\em IEEE Conf. Computer Vision and Pattern Recognition}, pages
  4511--4520, 2015.

\end{thebibliography}
